\documentclass{bmvc2k}

\usepackage{amsmath}
\usepackage{amssymb}
\usepackage{booktabs}
\usepackage[table]{xcolor}
\usepackage{graphicx}

\title{Semantic Slots for Video Object-Centric Learning} 

\addauthor{Khalil Sabri}{khalil.sabri@polymtl.ca}{1}
\addauthor{Guillaume-Alexandre Bilodeau}{gabilodeau@polymtl.ca}{1}
\addauthor{Nicolas Saunier}{nicolas.saunier@polymtl.ca}{1}
\addauthor{Wassim Bouachir}{wassim.bouachir@teluq.ca}{2}

\addinstitution{
Polytechnique Montr\'eal\\
Montr\'eal, Canada
}

\addinstitution{
Université TÉLUQ\\
Montr\'eal, Canada
}

\runninghead{Sabri et al.}{Semantic Slots for Video Object-Centric Learning}

\def\etal{\emph{et al}\bmvaOneDot}

\begin{document}

\maketitle

\begin{abstract}
Video Object-Centric Learning (OCL) has traditionally focused on refining the encoder architecture to ensure temporal consistency. In this paper, we argue that the primary bottleneck lies in the decoder. We show that traditional decoders force slots to be spatially anchored, hindering their ability to adapt to motion. We propose \textbf{SemanticSlots}, which uses a Transformer-based decoder that leverages image context, relieving slots from encoding boundary precision and spatial location. This allows slots to function as semantic queries that are inherently object position invariant, retrieving matching features rather than memorizing coordinates. More importantly, this property allows slots computed from a single frame to decompose subsequent video frames, eliminating the need for complex temporal predictors or auxiliary temporal losses. Results on YouTube-VIS show that SemanticSlots improves upon VideoSAUR by 31 points in mBO and outperforms current state-of-the-art methods by 21 points, achieving 86.6\% ARI and 62.8\% mBO. Our code and pre-trained models will be available at \url{https://github.com/sabrikhalil/Semantic-Slots}.
\end{abstract}

\section{Introduction}

Video Object-Centric Learning (OCL) aims to decompose videos into discrete object representations called slots~\cite{locatello2020object,kipf2021conditional}, learning to discover and segment objects across frames in an unsupervised manner. This structured representation is essential for downstream tasks such as visual reasoning, planning, and control~\cite{webb2023systematic,zadaianchuk2020self}. Prior work has largely prioritized the development of complex temporal objectives and encoder-side consistency~\cite{zadaianchuk2023object,manasyan2025temporally,zhao2025predicting}. We argue that this focus overlooks a critical structural bottleneck: the feature-agnostic nature of traditional decoders. \emph{Decoders that lack direct access to image features overburden low-capacity slot vectors, forcing them to explicitly encode absolute spatial coordinates and high-frequency textural details to satisfy the reconstruction objective}.

\begin{figure}[t!]
    \centering
    \includegraphics[width= 0.7 \linewidth]{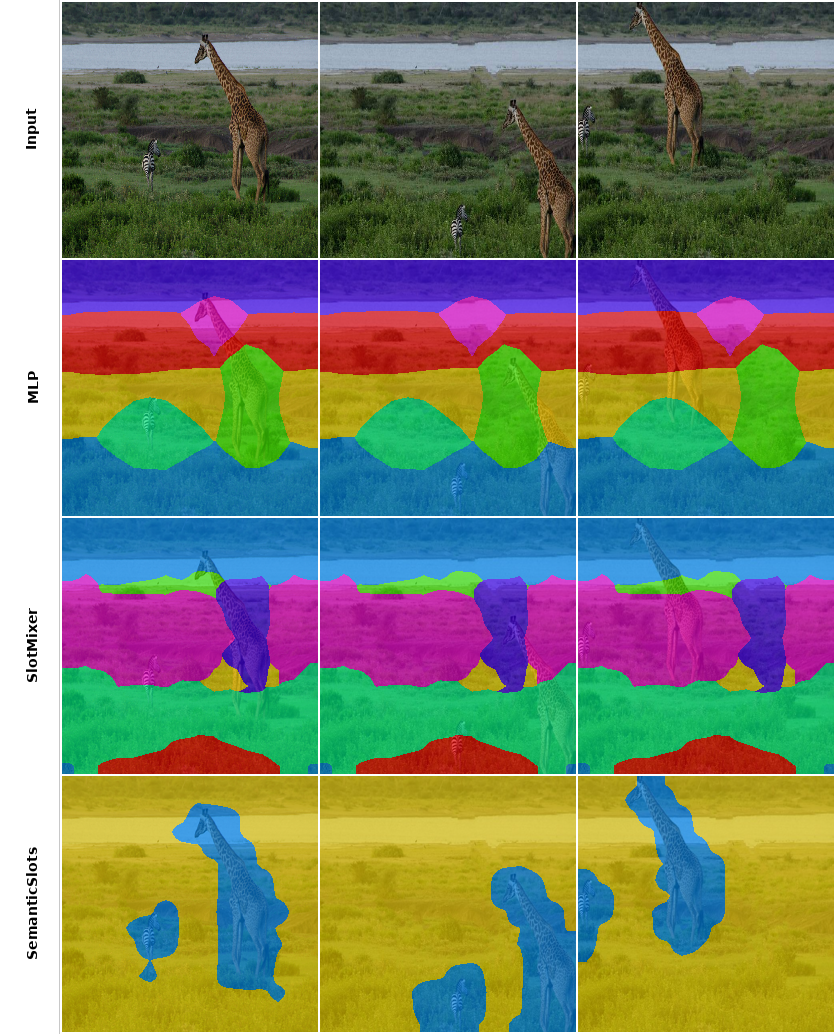}
    \caption{\textbf{Position Invariance via Context-Aware Decoding.} We test whether slots computed from one frame can decompose a spatially shifted version of that frame. 
    \textbf{Row 1:} Original frame and translated version. 
    \textbf{Row 2:} MLP decoder~\cite{seitzer2022bridging}. 
    \textbf{Row 3:} SlotMixer decoder~\cite{sajjadi2022object}. 
    \textbf{Row 4:} SemanticSlots-Frozen (ours). 
    Context-free decoders memorize the original positions and fail under displacement; our context-aware decoder re-localizes the object using the shifted feature map as the search space.}
    \label{fig:teaser}
\end{figure}

This fundamental limitation is visually demonstrated in Figure~\ref{fig:teaser}. As shown, traditional context-free decoders (MLP~\cite{seitzer2022bridging}, SlotMixer~\cite{sajjadi2022object}) structurally lack access to the new image context. Because they reconstruct features solely from slots, they are forced to engage in rigid memorization, reconstructing objects at their original spatial locations regardless of the input shift. In contrast, \emph{by equipping the decoder with cross-attention to the new image features, our model successfully re-localizes the object}. Even using the slots from the original image, the decoder uses the updated feature map as a search space, retrieving the object at its new location. This facilitates a shift from memorization to contextual retrieval, freeing slots from the need to encode spatial precision.

During the design of our method, we discovered that context-aware decoders~\citep{seitzer2022bridging}, originally designed for static images, induce an unexpected property: \textbf{slots become position-invariant}. We exploit this insight to propose \textbf{SemanticSlots}, which can compute slots once from the first frame and reuses them across the entire video. In our experiments, we demonstrate that even with these frozen slots, the cross-attention of the decoder re-binds the same semantic queries to objects in every subsequent frame, using the frame features as the search space. This shows that maintaining consistent object decomposition across a video does not require heavy temporal prediction or recurrent sequence modeling; it is a direct consequence of position invariance. We also show that refreshing these queries at each frame further sharpens attention maps, handles shape changes, and captures new objects that enter the video late, further improving performance. 

This work makes the following contributions:
\begin{itemize}

\item We show that context-aware decoding relieves slots from the burden of encoding spatial coordinates. This makes slots position-invariant, allowing slots computed on a single frame to segment the same object in subsequent frames;

\item We exploit this property to propose \textbf{SemanticSlots}, a video Object-Centric Learning framework trained entirely on single images without temporal supervision. We demonstrate that reusing initial slots as fixed semantic queries is sufficient to outperform recent state-of-the-art methods, completely eliminating complex temporal predictors and auxiliary losses;

\item We achieve state-of-the-art on YouTube-VIS (86.6\% ARI, 62.8\% mBO), outperforming the previous best by 39.8 and 21.3 points, through simplification, not added complexity.

\end{itemize}

\section{Related Work}

\subsection{Foundations and Feature-Based Object Discovery.}
Object-centric learning (OCL) decomposes images into discrete latent entities termed slots \cite{burgess2019monet,greff2019multi}. The Slot Attention \cite{locatello2020object} paradigm utilizes an iterative competitive bottleneck to extract representations for downstream relational reasoning \cite{webb2023systematic}, control \cite{zadaianchuk2020self}, and VQA \cite{didolkar2025ctrlo,yi2019clevrer}. While early models relied on pixel-level reconstruction \cite{locatello2020object,burgess2019monet,greff2019multi}, capacity was often wasted on high-frequency noise, failing to scale to complex scenes. DINOSAUR \cite{seitzer2022bridging} addressed this by reconstructing rich self-supervised foundation features from DINO \cite{caron2021emerging}, enabling object discovery in real-world images.

\subsection{Decoder Architectures and the Spatial Bottleneck.}
The decoder architecture directly shapes what information slots must encode. Traditional OCL models largely utilize context-free decoders, such as the Spatial Broadcast Decoder \cite{watters2019spatial} or MLP-based SlotMixers \cite{sajjadi2022object}, which lack direct access to image context during reconstruction and force slots to explicitly encode absolute spatial coordinates. \citet{singh2021illiterate} and \cite{singh2022simple} introduced autoregressive transformer decoders to handle complex visual dependencies by reconstructing discrete VQ-VAE tokens \cite{van2017neural}. Building on this, \citet{seitzer2022bridging} utilized a transformer-based decoder to reconstruct continuous self-supervised features, demonstrating superior decomposition in real-world scenes. However, high-capacity decoders can face training stability issues and may neglect slot inputs in favor of past ground-truth tokens \cite{singh2021illiterate,singh2022simple,seitzer2022bridging}. \citet{kakogeorgiou2024spot} addressed this by introducing patch-order permutations to ensure more reliance on slot vectors during decoding. While these advancements improved static image decomposition, prior work has not asked whether the resulting slots transfer across images. We make this transferability central, showing that such slots are position-invariant and exploiting this property to eliminate the temporal machinery of video Object-Centric Learning.

\subsection{Video Object Discovery.}
The temporal extension of OCL has primarily focused on the transitioner paradigm, utilizing recurrent slot updates to model spatiotemporal dynamics. SAVi and SAVi++ \cite{kipf2021conditional,elsayed2022savipp} established predictor-corrector frameworks that often rely on auxiliary motion or depth signals to handle complex real-world videos. Adopting the feature-reconstruction logic of DINOSAUR, VideoSAUR \cite{zadaianchuk2023object} achieved unsupervised scaling by employing a SlotMixer decoder alongside temporal similarity losses. Subsequent methods have focused on enhancing the encoder to better handle temporal consistency. \citet{manasyan2025temporally} introduced a contrastive loss between slots across frames to enforce object identity preservation, while \citet{zhao2025predicting} employed a learned query predictor that leverages future frames to guide the transitioner in modeling object dynamics. We observe that these approaches over-burden low-capacity slot vectors with the dual task of representing semantic identity and absolute spatial coordinates across time. In contrast, our work offloads the task of spatial localization to a context-aware decoder, allowing slots to function as stable semantic queries that maintain identity across frames through cross-attention.

\section{Methodology}

\begin{figure}[t!]
    \centering
    \includegraphics[width=\textwidth]{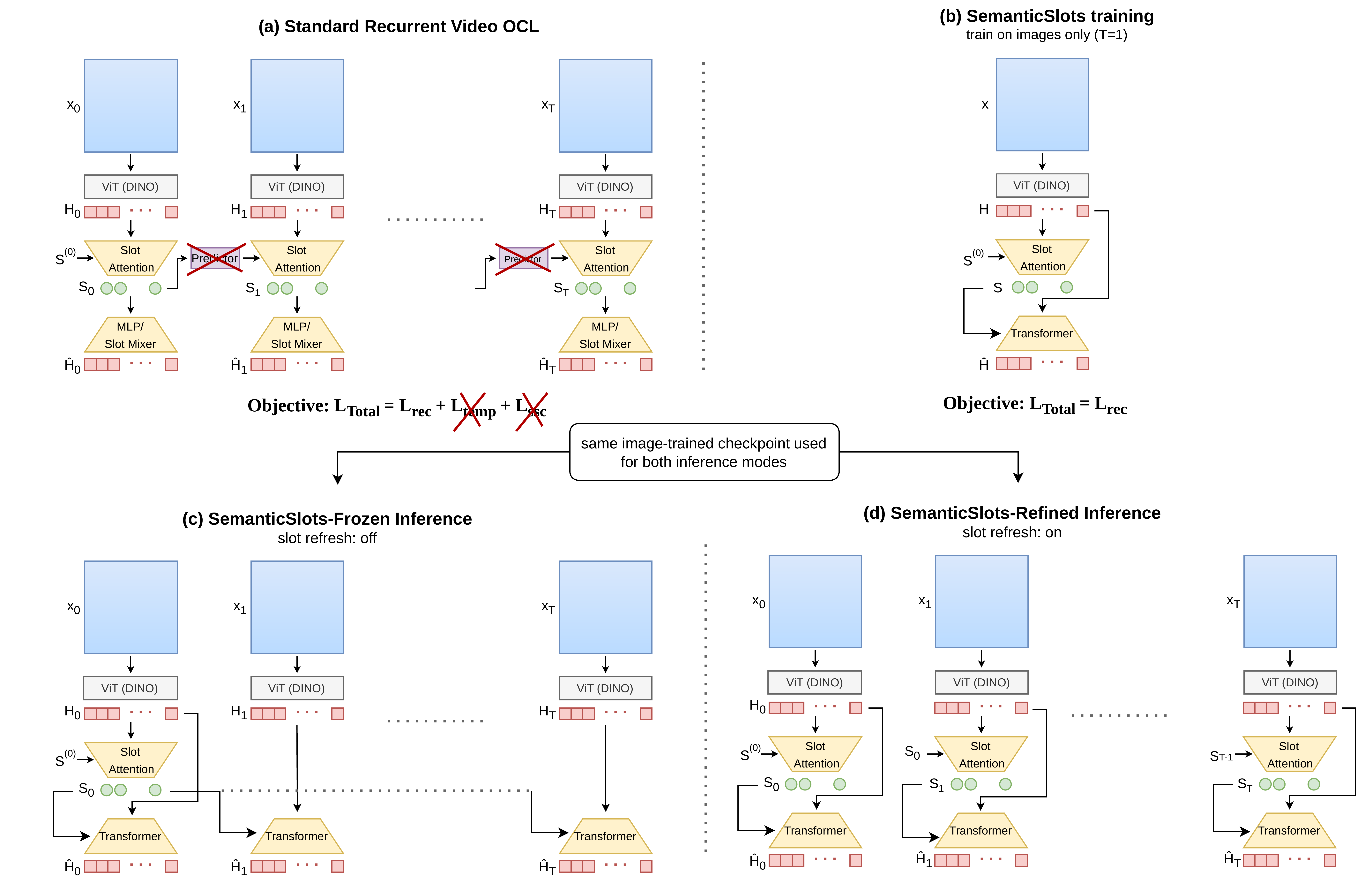}
    \caption{
Architecture motivation and overview of SemanticSlots. 
(a) Standard recurrent video OCL maintains temporal consistency through recurrent slot updates, predictors, and temporal losses. 
(b) SemanticSlots is trained only on static images using a reconstruction loss. 
(c) SemanticSlots-Frozen computes slots once from the first frame and reuses them as fixed semantic queries for all later frames. 
(d) SemanticSlots-Refined uses the same image-trained checkpoint, but refreshes slots at each frame during inference, without predictors, temporal losses, or video-level training.
}
\label{fig:architecture}
\end{figure}

We introduce the foundations of OCL and extend the explanations to the video setting (Sec.~\ref{sec:preliminaries}). We then identify the spatial entanglement problem in standard decoders (Sec.~\ref{sec:entanglement}) and present SemanticSlots, our solution for position-invariant object discovery (Sec.~\ref{sec:SemanticSlots}).

\subsection{Preliminaries}
\label{sec:preliminaries}

\paragraph{Object-Centric Learning.}
Object-Centric Learning (OCL) decomposes images into discrete latent entities called \emph{slots}~\cite{locatello2020object}. Each slot is a $D$-dimensional vector representing a single object or scene component. Given $K$ slots, the slot bank is denoted $S = \{s_1, \ldots, s_K\} \in \mathbb{R}^{K \times D}$.

Modern OCL methods reconstruct dense patch features rather than raw pixels~\cite{seitzer2022bridging}, avoiding wasted model capacity on high-frequency textural details and allowing slots to focus on semantic object properties. Given an input image $x \in \mathbb{R}^{H \times W \times 3}$, a pretrained Vision Transformer encoder, typically DINO or DINOv2~\cite{caron2021emerging,oquab2023dinov2}, extracts $N$ patch features, where the patches are flattened into a sequence following a standard raster-scan order (from top-left to bottom-right):
\begin{equation}
    H = \text{ViT}(x), \quad H = \{h_1, \ldots, h_N\}, 
\end{equation}
where $h_i \in \mathbb{R}^{D_f}$ is the feature vector for patch $i$, and $N$ is determined by the patch size (e.g., $N{=}196$ for $224{\times}224$ images with $16{\times}16$ patches).

Slot Attention~\cite{locatello2020object} is a competitive binding mechanism that groups patch features into slots. Slots are initialized from a learned multidimensional Gaussian distribution $S^{(0)} \sim \mathcal{N}(\mu, \sigma)$ and refined through $R$ iterations of attention, where slots compete via softmax normalization over the slot dimension. After refinement, each slot $s_k$ aggregates features from patches that share similar visual properties, ideally corresponding to distinct objects:
\begin{equation}
    S = \text{SlotAttn}(H, S^{(0)}).
\end{equation}

Then, a decoder typically reconstructs patch features from slots. In context-free decoders (e.g., MLPs, SlotMixers), the output depends only on the slots:
\begin{equation}
    \hat{H} = \text{Decoder}(S), \quad \hat{H} = \{\hat{h}_1, \ldots, \hat{h}_N\}.
    \label{eq:context_free_decoder}
\end{equation}
Training minimizes the reconstruction loss between predicted and target ViT features:
\begin{equation}
    \mathcal{L}_{\text{recon}} = \|H - \hat{H}\|^2.
\end{equation}

\paragraph{Video Object-Centric Learning.}
Extending OCL to video requires maintaining object identity across frames. Given a video $\{x_0, x_1, \ldots, x_T\}$, standard methods~\cite{kipf2021conditional,zadaianchuk2023object} process each frame $t$ sequentially as shown in Figure~\ref{fig:architecture} (Left):
\begin{align}
    H_t &= \text{ViT}(x_t), \\
    S_t &= \text{SlotAttn}(H_t, S_{t-1}), \\
    \hat{H}_t &= \text{Decoder}(S_t).
\end{align}
Slot Attention at frame $t$ is initialized with the slots from the previous frame $S_{t-1}$, not from random noise. This recurrent initialization provides temporal continuity. For $t{=}0$, slots are sampled from the learned Gaussian: $S_{-1} = S^{(0)}$.

To further enforce temporal consistency, most methods introduce additional components: (1) a Predictor module (e.g., GRU or Transformer) that models slot dynamics between frames~\cite{kipf2021conditional,zhao2025predicting}, and (2) temporal losses such as feature similarity ($\mathcal{L}_{temp}$)~\cite{zadaianchuk2023object} or contrastive terms ($\mathcal{L}_{ssc}$)~\cite{manasyan2025temporally} that explicitly supervise slot correspondence across time. 

\subsection{Spatial Entanglement in Standard Decoders}
\label{sec:entanglement}

Standard OCL decoders, including MLPs~\cite{seitzer2022bridging} and SlotMixers~\cite{sajjadi2022object}, are \emph{context-free}: they reconstruct the full set of patch features $\hat{H}$ solely from slots as defined in equation~\eqref{eq:context_free_decoder}, without access to the current image features.

This creates a fundamental problem: to minimize reconstruction error, slots must encode \emph{both semantic identity (what the object looks like) and absolute spatial coordinates (where the object is)}. We term this \textbf{spatial entanglement}. Because slots are anchored to specific positions, they cannot adapt when objects move, the decoder reconstructs objects at memorized locations regardless of actual motion.

Figure~\ref{fig:teaser} demonstrates this limitation. When we freeze slots from an original image and decode against a spatially-shifted version, context-free decoders fail catastrophically, reconstructing at the original position instead of following the displacement.

\subsection{SemanticSlots: Position-Invariant Object Discovery}
\label{sec:SemanticSlots}

We present \textbf{SemanticSlots}, our approach for video object discovery that eliminates the need for temporal predictors and auxiliary losses.

We adopt the Transformer-based decoder design of \citet{seitzer2022bridging} and \citet{vaswani2017attention} that reconstructs features autoregressively over the spatial patch sequence within each frame. Our contribution is not the decoder architecture itself, but the discovery that it produces position-invariant slots, a property we exploit to apply slots from one frame to entirely different frames at inference time. Unlike context-free decoders (Sec.~\ref{sec:entanglement}), each patch prediction $\hat{h}_{t,i}$ at frame $t$ is conditioned on both the slots $S_0$ (computed once from frame 0) and the preceding features from the same frame $H_{t,<i} = \{h_{t,1}, \ldots, h_{t,i-1}\}$:

\begin{equation}
    \hat{h}_{t,i} = \text{TransformerDecoder}(S_0, H_{t,<i}).
\end{equation}
The decoder uses cross-attention to retrieve information from slots, and causal self-attention over $H_{t,<i}$ to leverage spatial context within frame $t$. This architectural choice shifts precise spatial encoding from slots to the decoder, slots no longer need to memorize coordinates, as the decoder can infer spatial structure from the feature context.

Our key insight is that context-aware decoding enables simple video inference without predictors or temporal losses, as illustrated in Figure~\ref{fig:architecture}. In the strict frozen setting, slots computed from the first frame can be reused as semantic queries for later frames. Given a video $\{x_0, x_1, \ldots, x_T\}$:

\begin{align}
    H_t &= \text{ViT}(x_t), \quad \forall t \in \{0, \ldots, T\}, \\
    S_0 &= \text{SlotAttn}(H_0, S^{(0)}), \\
    \hat{H}_t &= \text{TransformerDecoder}(S_0, H_t), \quad \forall t \in \{0, \ldots, T\}.
\end{align}
Hence, we compute slots \emph{once} at $t{=}0$ and reuse them for all subsequent frames. This is what we term \textbf{frozen slots}, the slot bank $S_0$ remains fixed throughout the video. Based on this pipeline, we evaluate two modes for our framework:
(1) \textbf{SemanticSlots-Frozen}, where slots remain strictly fixed after the initial frame as presented in the equations above, and (2) \textbf{SemanticSlots-Refined}, where standard slot attention iterations are applied at each frame. While the frozen mode serves as a powerful baseline, we find that keeping the frame-wise slot attention updates used in classic OCL methods consistently improves results. This simple step successfully handles dynamic variations, such as objects entering or leaving the scene, or sudden scale changes, providing a lightweight solution to these common edge cases.

\paragraph{Segmentation via Cross-Attention.}
At inference, the reconstructed features $\hat{H}_t$ are completely discarded. The reconstruction task is only used during training to force the model to learn slots. The actual output of our method, the object segmentation, is obtained directly from the decoder cross-attention weights, which assign each patch to a specific slot.

\paragraph{Contrast with Standard Video OCL.}
Our approach differs fundamentally from existing video object-centric methods. VideoSAUR~\cite{zadaianchuk2023object}, SlotContrast~\cite{manasyan2025temporally}, and RandSF.Q~\cite{zhao2025predicting} all rely on: (1) \emph{predictors} (GRU or Transformer modules) that model slot dynamics between frames, (2) \emph{temporal losses} (feature similarity~\cite{zadaianchuk2023object} or contrastive terms~\cite{manasyan2025temporally}) that explicitly supervise slot correspondence across time, and (3) complex \emph{video-level training pipelines}. SemanticSlots requires neither predictors nor temporal losses, and trains entirely on single static images. In our framework, frame-wise iterations are used only as an inference-time slot refresh, not as part of a video-trained recurrent pipeline.

\paragraph{Spatial Invariance.}
How can fixed slots remain bound to moving objects? The answer lies in the cross-attention mechanism. When decoding frame $t$, the decoder cross-attends to $H_t$. A slot $s_k$ attends to patches in $H_t$ that are semantically similar to the object it represents. If the object moves between frames, the attention pattern simply shifts to the new location, the slot content (the query) remains unchanged, only the attention target changes. This makes slots inherently position-invariant: \emph{they encode what an object is, while the decoder attention mechanism resolves where it appears in each frame}.

\paragraph{Training.}
We train exclusively on static images, as shown in Figure~\ref{fig:architecture}(b), using the standard reconstruction objective:
\begin{equation}
    \mathcal{L} = \|H - \hat{H}\|^2.
\end{equation}
Despite never observing video during training, the position-invariant property transfers directly to the temporal domain: slots learned on single images generalize to decomposing every frame of a video at test time.

\section{Experiments}
\subsection{Experimental Setup}

\paragraph{Datasets.} Following recent standards in video OCL \cite{manasyan2025temporally,zhao2025predicting}, we evaluate on both synthetic and real-world benchmarks. For synthetic data, we use MOVi-C and MOVi-D \cite{greff2022kubric}, which feature objects with complex textures and chaotic dynamics. For real-world evaluation, we utilize the high-quality version of YouTube-VIS (YTVIS) \cite{yang2019video}, with diverse natural videos.

\paragraph{Compared State-of-the-Art (SOTA) methods.} We compare against the primary lineage of video Slot Attention models: STEVE \cite{singh2022simple}, an early video-based Slot Attention model; VideoSAUR \cite{zadaianchuk2023object}, the first to scale to real-world features; SlotContrast \cite{manasyan2025temporally}, the previous state-of-the-art for temporal consistency; and RandSF.Q \cite{zhao2025predicting}, the most recent transition-based SOTA. We exclude methods like SAVi \cite{kipf2021conditional} or SOLV \cite{aydemir2023self} to ensure a fair comparison, as they rely on additional supervision (flow/depth) or specialized slot-pruning heuristics.

\paragraph{Metrics.} To evaluate the quality and temporal consistency of the discovered objects, we utilize a suite of recognized video-centric metrics. Following the evaluation protocol in \citet{zadaianchuk2023object,manasyan2025temporally}, we report the Adjusted Rand Index (ARI) to assess overall scene decomposition and the foreground ARI (ARI$_{fg}$) to measure the semantic grouping of foreground entities. To assess spatial precision and mask sharpness, we use mean Best Overlap (mBO) and mean Intersection over Union (mIoU), the latter serving as our strictest measure of segmentation accuracy.

All metrics are computed globally over the full video sequence to reflect the model ability to maintain stable object identities and achieve consistent object decomposition. Segmentation masks are generated by binarizing the decoder cross-attention maps along the slot dimension, which are then compared directly against ground-truth video annotations to quantify decomposition consistency. This protocol ensures that we measure how well fixed semantic queries decompose every frame of the video, not just the first.

\paragraph{Implementation Details.}
For YTVIS, we utilize a frozen ViT-B/14 DINOv2 backbone~\cite{oquab2023dinov2}. For MOVi-C and MOVi-D, we use ViT-B/14 DINOv1~\cite{caron2021emerging}. Consistent with observations in~\citet{manasyan2025temporally}, we found DINOv2 features led to training instability on these synthetic benchmarks.  
The Transformer decoder is configured with 4 layers and 4 attention heads with a hidden dimension of 768. We employ a batch size of 64 and a learning rate of \textbf{$2\times10^{-4}$}. All results report mean and standard deviation over three independent runs. We use K=7 for YTVIS, K=11 for MOVi-C, and K=21 for MOVi-D, following the standard settings in \citet{zadaianchuk2023object} and \citet{zhao2025predicting} for a fair comparison. Unless otherwise specified, all qualitative analyses and ablation studies are conducted on the YTVIS dataset.

\subsection{Comparison with State-of-the-Art}

\begin{table}[t!]
\centering
\caption{Comparison with state-of-the-art video object discovery. \textbf{Bold}: best result. \underline{Underline}: second best. Higher is better. Baseline results are from Zhao \etal~\cite{zhao2025predicting}}
\label{tab:main_results}
\small
\setlength{\tabcolsep}{1.5pt}

\vspace{0.25cm} 

\begin{tabular}{l cccc cccc}
\toprule
 & \multicolumn{4}{c}{MOVi-C (\#slot=11)} & \multicolumn{4}{c}{MOVi-D (\#slot=21)} \\
\cmidrule(lr){2-5} \cmidrule(lr){6-9}
\textbf{Method} & \textbf{ARI} & \textbf{ARI$_{fg}$} & \textbf{mBO} & \textbf{mIoU} & \textbf{ARI} & \textbf{ARI$_{fg}$} & \textbf{mBO} & \textbf{mIoU} \\ 
\midrule
STEVE & -- & -- & -- & -- & 32.7{\tiny$\pm$0.2} & 66.5{\tiny$\pm$0.2} & 23.0{\tiny$\pm$0.3} & 21.2{\tiny$\pm$0.3} \\
VideoSAUR & 41.9{\tiny$\pm$1.1} & 53.3{\tiny$\pm$2.1} & 16.1{\tiny$\pm$0.4} & 14.8{\tiny$\pm$0.4} &  -- & -- & -- & --  \\
SlotContrast & 64.6{\tiny$\pm$9.4} & 59.9{\tiny$\pm$5.3} & 27.7{\tiny$\pm$3.0} & 25.8{\tiny$\pm$2.9} & \underline{45.3}{\tiny$\pm$4.1} & 63.9{\tiny$\pm$0.2} & 26.7{\tiny$\pm$1.0} & 25.1{\tiny$\pm$0.9} \\
RandSF.Q (tsim) & 64.0{\tiny$\pm$2.9} & 66.3{\tiny$\pm$1.7} & 28.4{\tiny$\pm$1.3} & 26.1{\tiny$\pm$1.1} & 41.2{\tiny$\pm$2.2} & 72.0{\tiny$\pm$1.1} & 27.1{\tiny$\pm$0.9} & 25.4{\tiny$\pm$0.9} \\
RandSF.Q (ssc) & \underline{65.4}{\tiny$\pm$10.7} & \underline{67.4}{\tiny$\pm$2.1} & \underline{29.2}{\tiny$\pm$3.8} & \underline{26.8}{\tiny$\pm$3.7} & 41.6{\tiny$\pm$3.7} & \textbf{77.5}{\tiny$\pm$1.0} & \underline{27.4}{\tiny$\pm$1.0} & \underline{25.6}{\tiny$\pm$1.0} \\
\midrule
\midrule
\textbf{SemanticSlots} & \textbf{84.0}{\tiny$\pm$2.7} & \textbf{66.8}{\tiny$\pm$5.2} & \textbf{36.4}{\tiny$\pm$0.8} & \textbf{34.5}{\tiny$\pm$0.8} & \textbf{62.7}{\tiny$\pm$8.5} & \underline{76.4}{\tiny$\pm$3.0} & \textbf{35.5}{\tiny$\pm$3.0} & \textbf{33.9}{\tiny$\pm$2.7} \\ 
\bottomrule
\end{tabular}

\vspace{0.25cm} 

\begin{tabular}{l cccc}
\toprule
 & \multicolumn{4}{c}{YTVIS (\#slot=7)} \\
\cmidrule(lr){2-5}
\textbf{Method} & \textbf{ARI} & \textbf{ARI$_{fg}$} & \textbf{mBO} & \textbf{mIoU} \\ 
\midrule
STEVE & -- & -- & -- & -- \\
VideoSAUR & 34.4{\tiny$\pm$0.6} & 48.9{\tiny$\pm$1.5} & 31.4{\tiny$\pm$1.7} & 30.9{\tiny$\pm$0.3} \\
SlotContrast & 38.1{\tiny$\pm$0.7} & 48.8{\tiny$\pm$1.5} & 34.5{\tiny$\pm$0.3} & 34.4{\tiny$\pm$0.2} \\
RandSF.Q (tsim) & \underline{46.8}{\tiny$\pm$0.7} & \underline{60.7}{\tiny$\pm$1.7} & \underline{41.5}{\tiny$\pm$0.2} & \underline{40.6}{\tiny$\pm$0.1} \\
RandSF.Q (ssc) & 41.5{\tiny$\pm$0.1} & 58.9{\tiny$\pm$0.9} & 39.4{\tiny$\pm$0.4} & 39.0{\tiny$\pm$0.4} \\
\midrule
\textbf{SemanticSlots} & \textbf{86.6}{\tiny$\pm$0.3} & \textbf{77.5}{\tiny$\pm$0.6} & \textbf{62.8}{\tiny$\pm$0.4} & \textbf{60.5}{\tiny$\pm$0.5} \\
\bottomrule
\end{tabular}
\end{table}

Our main framework, SemanticSlots (evaluated in its default Refined mode) achieves SOTA results across nearly all evaluated metrics on both synthetic and real-world benchmarks (Table~\ref{tab:main_results}). On YouTube-VIS, we surpass the previous best transition-based method by 39.8 points in ARI and 21.3 points in mBO. We observe similarly significant gains on the challenging MOVi benchmarks under identical evaluation settings. On MOVi-C, our framework outperforms the strongest baseline by 18.1 points in ARI and 8.0 points in mBO. On MOVi-D, despite the high slot pressure ($K{=21}$), SemanticSlots successfully scales, establishing a new SOTA baseline of 62.7\% ARI and 35.5\% mBO. These results validate our hypothesis: a retrieval-based decoder mitigates the spatial-bias bottleneck in context-free decoders.

\begin{figure}[t]
    \centering
    \begin{tabular}{@{}c@{\hspace{5pt}}c@{}}
        \includegraphics[width=0.48\textwidth]{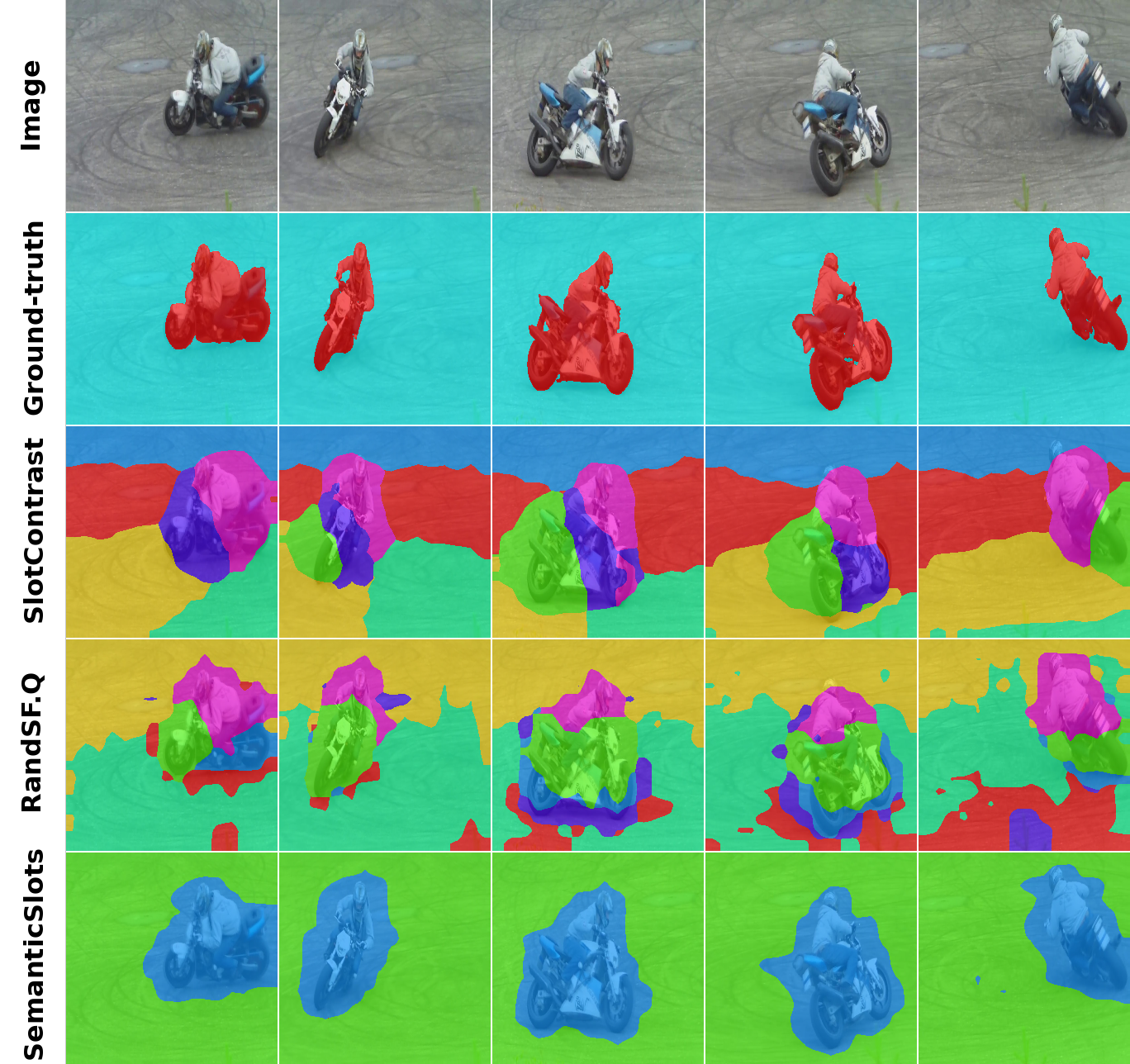} & 
        \includegraphics[width=0.48\textwidth]{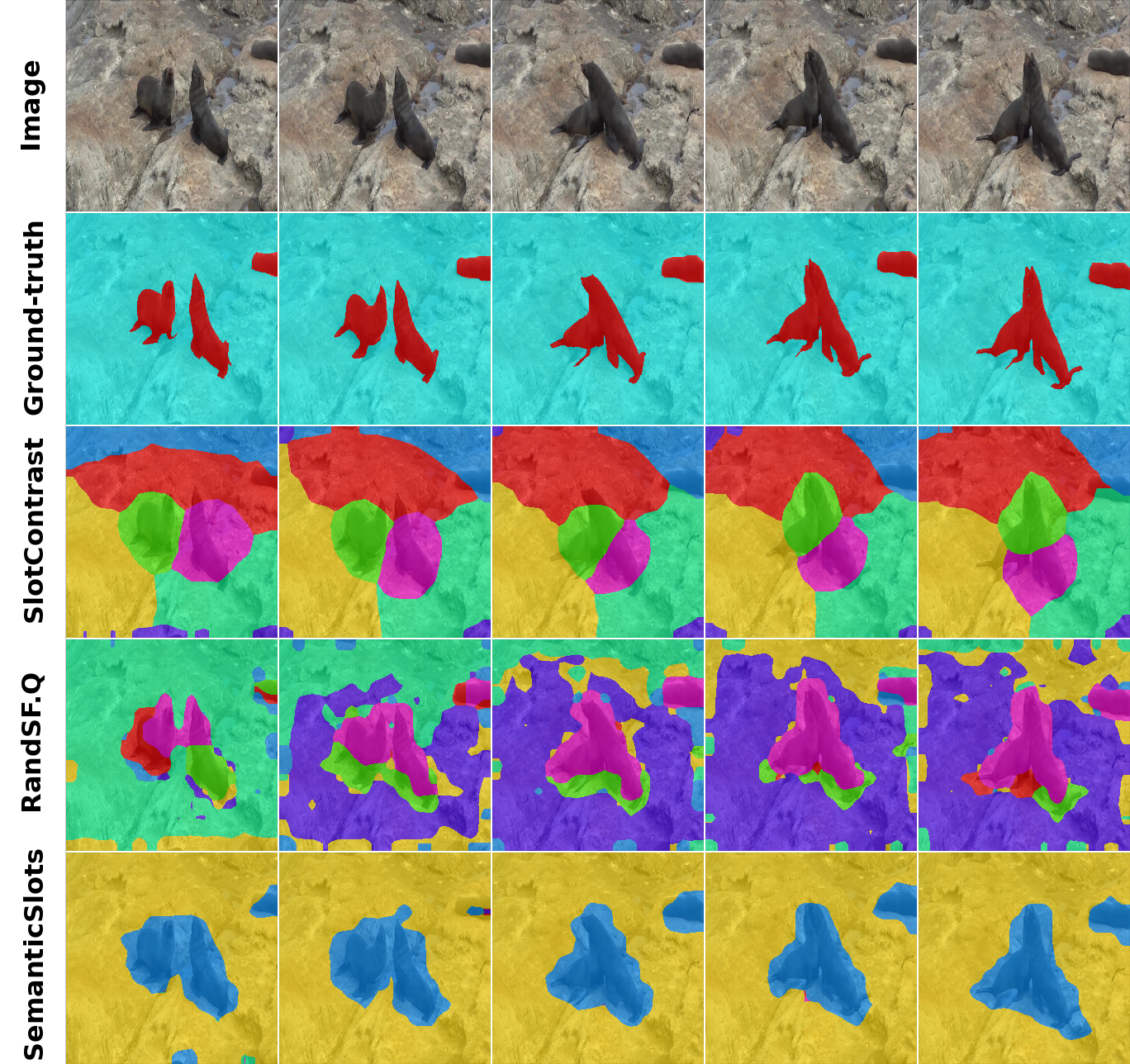}
    \end{tabular}
    \caption{\textbf{Qualitative comparison on two sequences from YouTube-VIS.} Rows from top to bottom represent: (1) Input frames, (2) Ground-truth masks, (3) SlotContrast \protect\cite{manasyan2025temporally}, (4) RandSF.Q \protect\cite{zhao2025predicting}, and (5) Our approach. Our method shows superior boundary precision and reduced classification noise compared to transition-based baselines.}
    \label{fig:qualitative_results}
\end{figure}
\begin{figure*}[p]
    \centering
    
    \vspace{-0.3cm}
    {\small \textbf{YouTube-VIS}} \\
    \vspace{0.05cm}
    \begin{tabular}{@{}c@{\hspace{5pt}}c@{}}
        \includegraphics[width=0.42\textwidth]{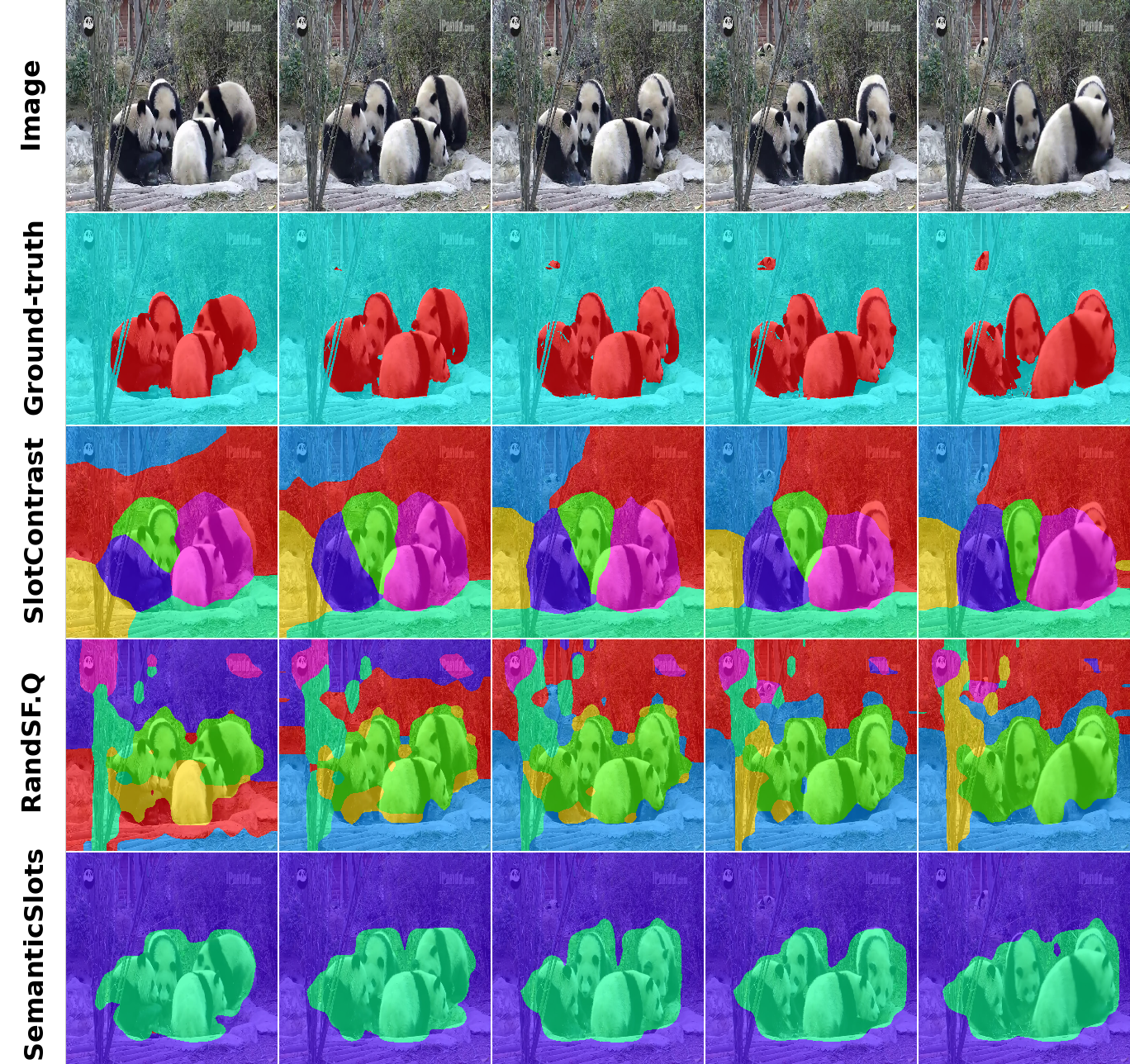} & 
        \includegraphics[width=0.42\textwidth]{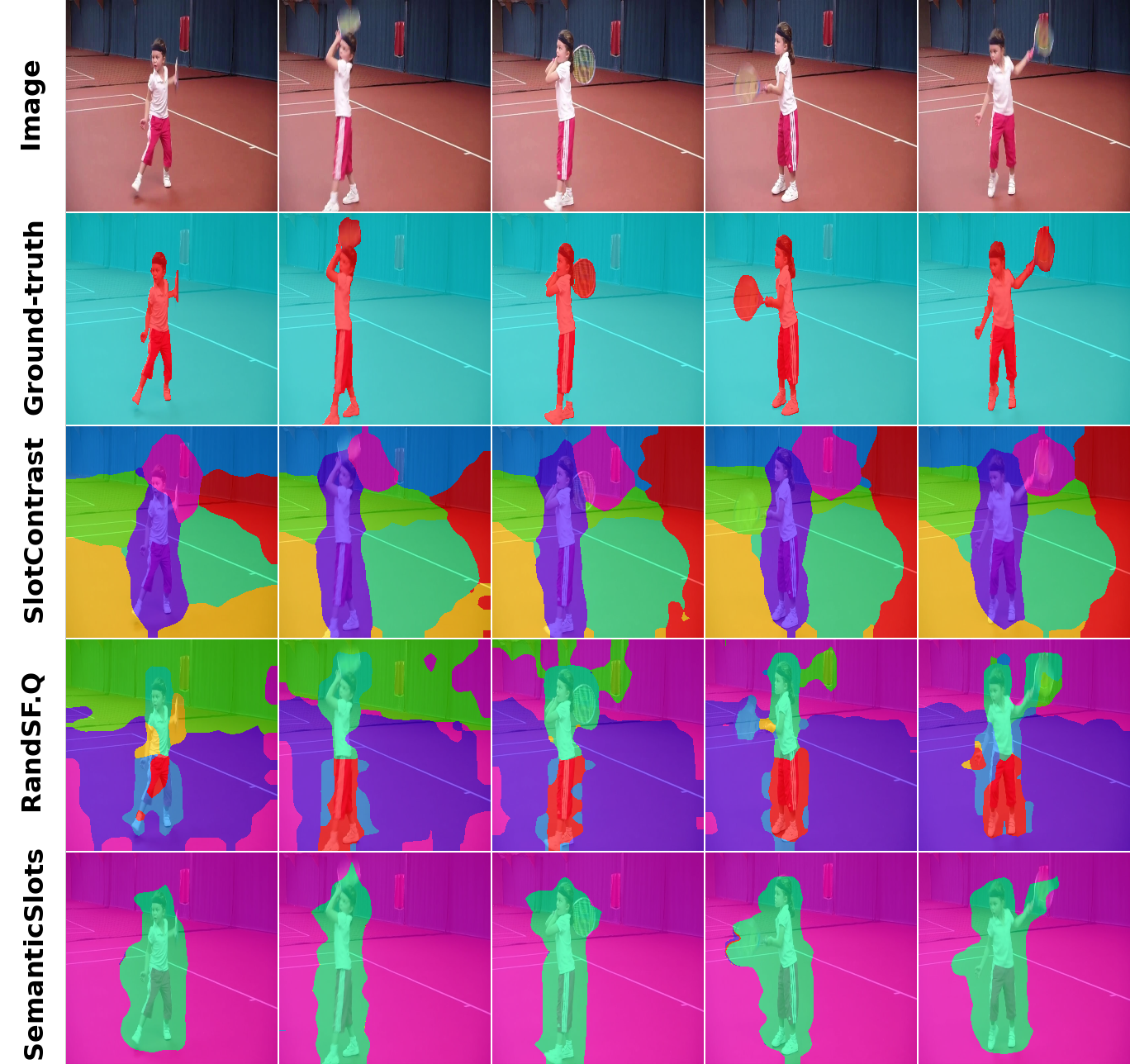}
    \end{tabular}
    
    \vspace{0.15cm}
    
    {\small \textbf{MOVi-C}} \\
    \vspace{0.05cm}
    \begin{tabular}{@{}c@{\hspace{5pt}}c@{}}
        \includegraphics[width=0.42\textwidth]{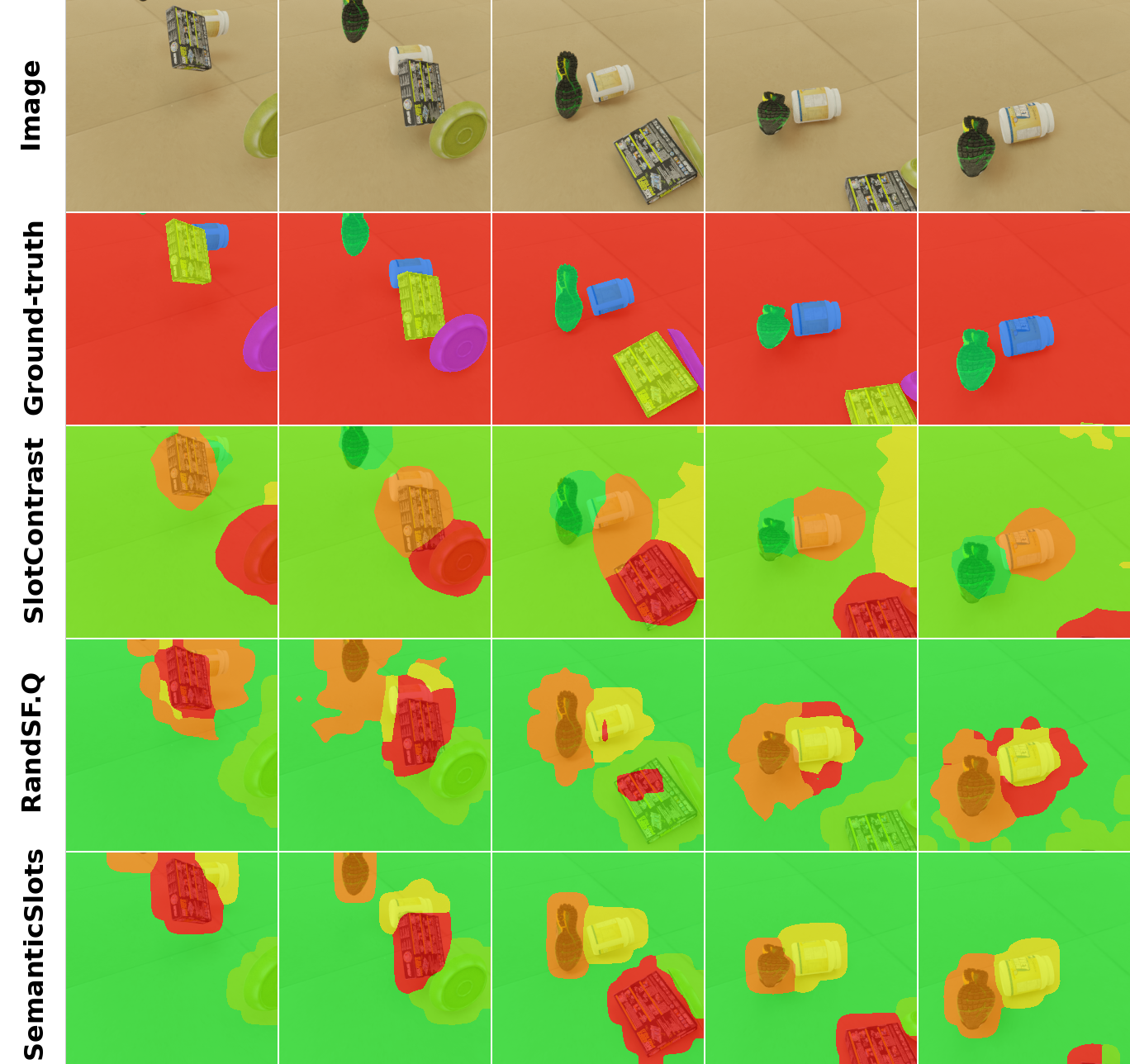} & 
        \includegraphics[width=0.42\textwidth]{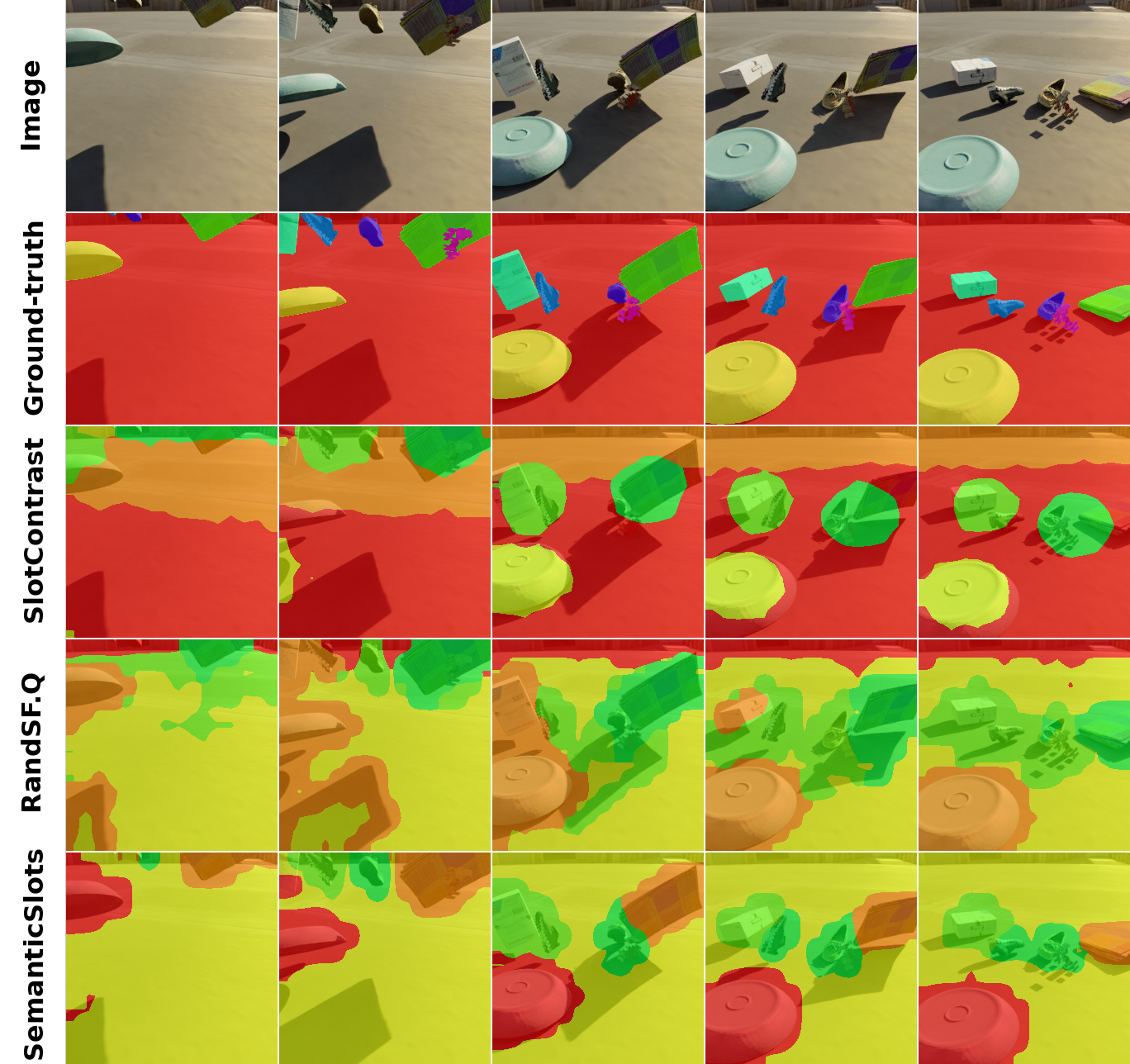}
    \end{tabular}
    
    \vspace{0.15cm}
    
    {\small \textbf{MOVi-D}} \\
    \vspace{0.05cm}
    \begin{tabular}{@{}c@{\hspace{5pt}}c@{}}
        \includegraphics[width=0.42\textwidth]{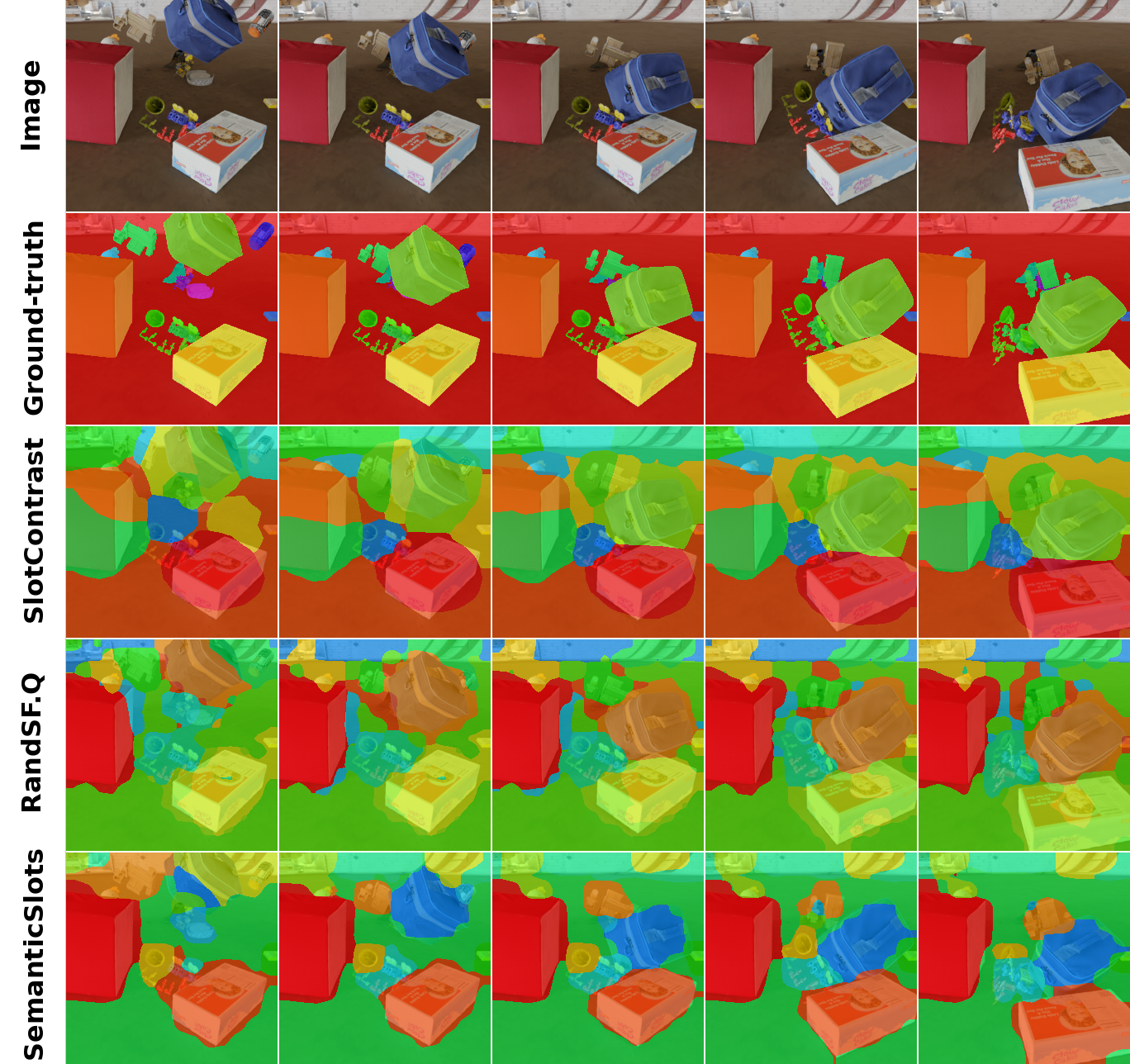} & 
        \includegraphics[width=0.42\textwidth]{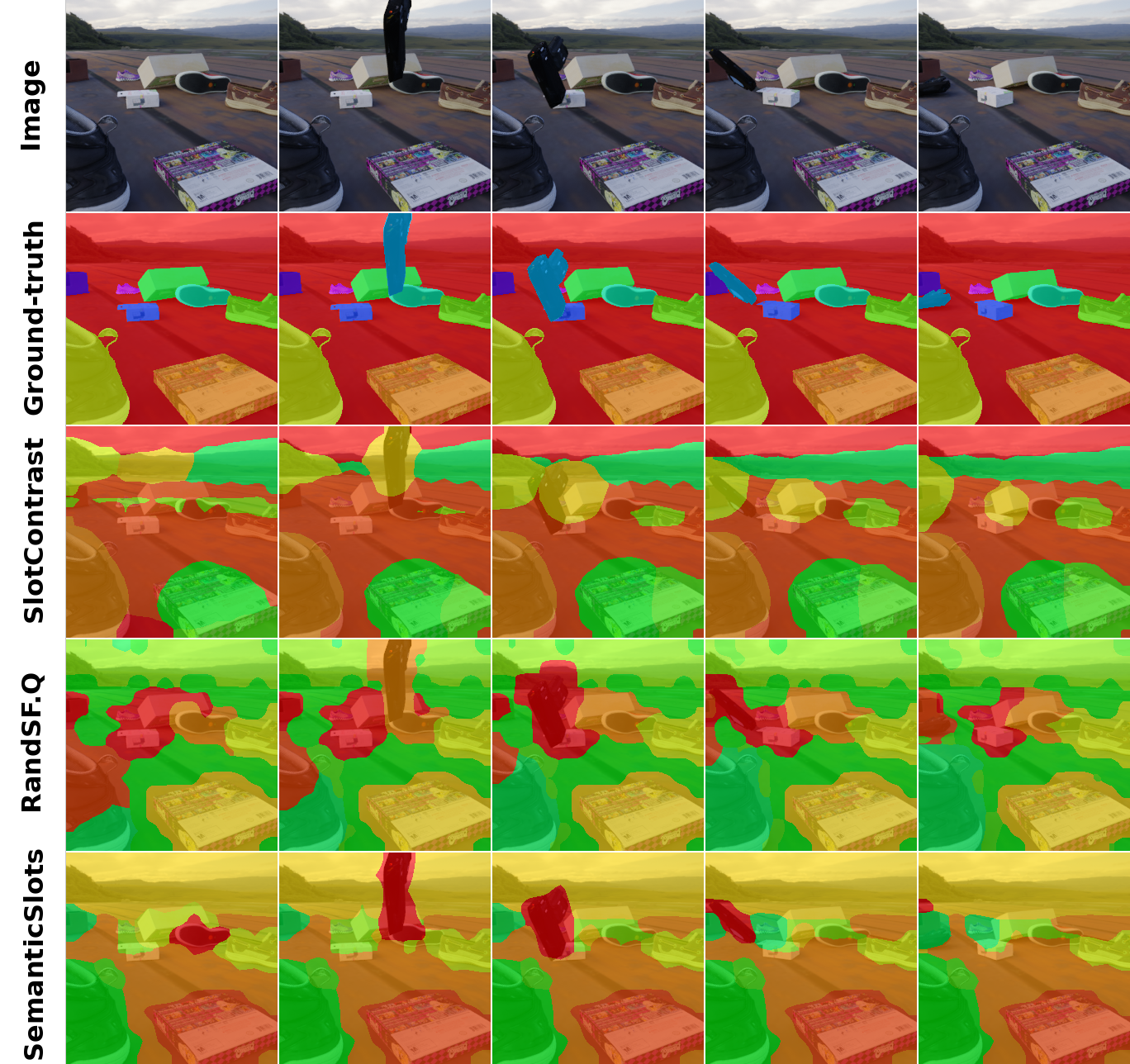}
    \end{tabular}
    
    \vspace{0.2cm}
    \caption{\textbf{Comprehensive qualitative comparison across diverse visual domains.} Rows within each sequence display, from top to bottom: (1) Input video frames, (2) Ground-truth instance masks, (3) SlotContrast~\cite{manasyan2025temporally}, (4) RandSF.Q~\cite{zhao2025predicting}, and (5) Our approach (SemanticSlots). Our framework yields significantly more stable identity assignments, tracks shapes with better boundary precision, and inherently eliminates the background over-clustering noise plaguing transition-based recurrent baselines.}
    \label{fig:full_page_qualitative}
\end{figure*}

We give examples of predicted masks for YouTube-VIS in Figure~\ref{fig:qualitative_results}. We observe that our approach generally yields more spatially consistent masks than the previous SOTA methods. As seen in the figure, methods such as SlotContrast and RandSF.Q occasionally exhibit flickering or misassigned patches across frames. We attribute our model stability to its architectural simplicity; by utilizing a retrieval-based decoder without a temporal predictor, we avoid the error propagation common in transition-based frameworks. Furthermore, this simplified objective facilitates a smoother training process, which we find leads to sharper object boundaries and more stable identity assignments throughout the sequence. For a more comprehensive visual analysis, we include extended qualitative results across the different datasets (YTVIS, MOVi-C and MOVi-D), please refer to Figure~\ref{fig:full_page_qualitative}.

Notably, SemanticSlots tends to use fewer active slots than the allocated $K$. This occurs because context-free decoders must encode all spatial and textural information in slots, bene\-fiting from more slots. In contrast, our Transformer decoder leverages image context, making additional slots redundant, fewer semantic queries suffice when the decoder can retrieve spatial details directly. This finding provides a numerical explanation for the significantly higher ARI achieved across all benchmarks; by requiring fewer slots to represent the scene, our model inherently avoids the background over-clustering common in SOTA methods.

\subsection{Further Analysis and Ablation Experiments}

\paragraph{Handling Dynamic Scenes and the Role of Slot Refreshing.}
SemanticSlots-Frozen assumes that the first frame already contains all relevant scene components. This works well when objects remain visible throughout the video, but can fail when new objects enter later or when objects are initially occluded. To study this failure mode, we evaluate SemanticSlots-Adaptive, a simple inference-time variant that refreshes slots only when the current frame is not well explained by the initial scene.

The trigger operates in DINO space. After running Slot Attention on the first frame, each slot $s_k$ is converted into a semantic anchor $a_k$ by averaging its attended patch features:
\begin{equation}
    a_k = \sum_i \alpha_{k,i} h_{0,i},
\end{equation}
where $\alpha_{k,i}$ is the attention weight between slot $k$ and patch $i$ in the initial frame. Intuitively, if a slot binds to the background, its anchor represents the background directly in DINO space.

For a later frame $t$, each patch feature $h_{t,i}$ is compared to its closest anchor:
\begin{equation}
    d_{t,i} = 1 - \max_k \cos(h_{t,i}, a_k).
\end{equation}

\begin{figure}[ht]
    \centering
    \includegraphics[width=0.48\linewidth]{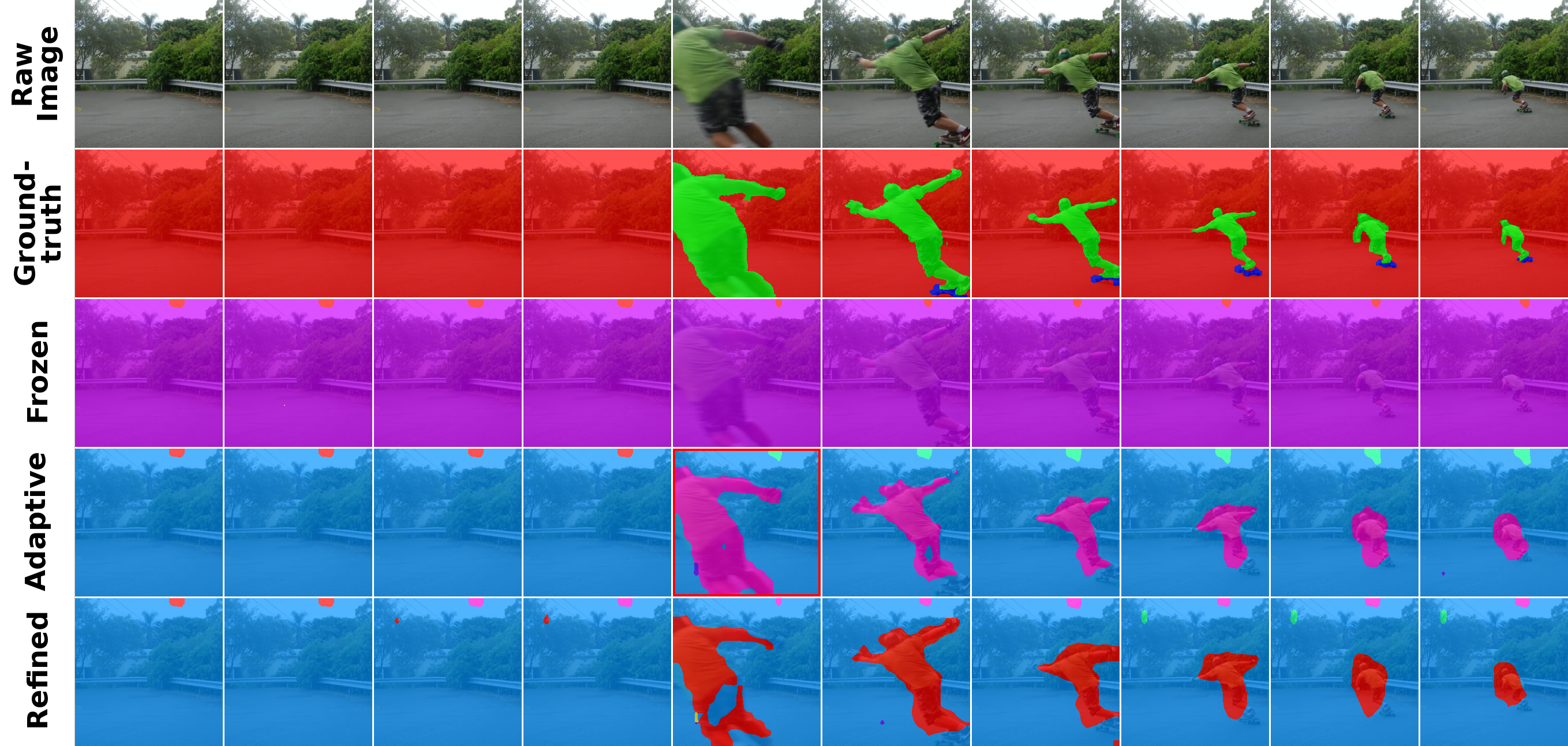} \hfill
    \includegraphics[width=0.48\linewidth]{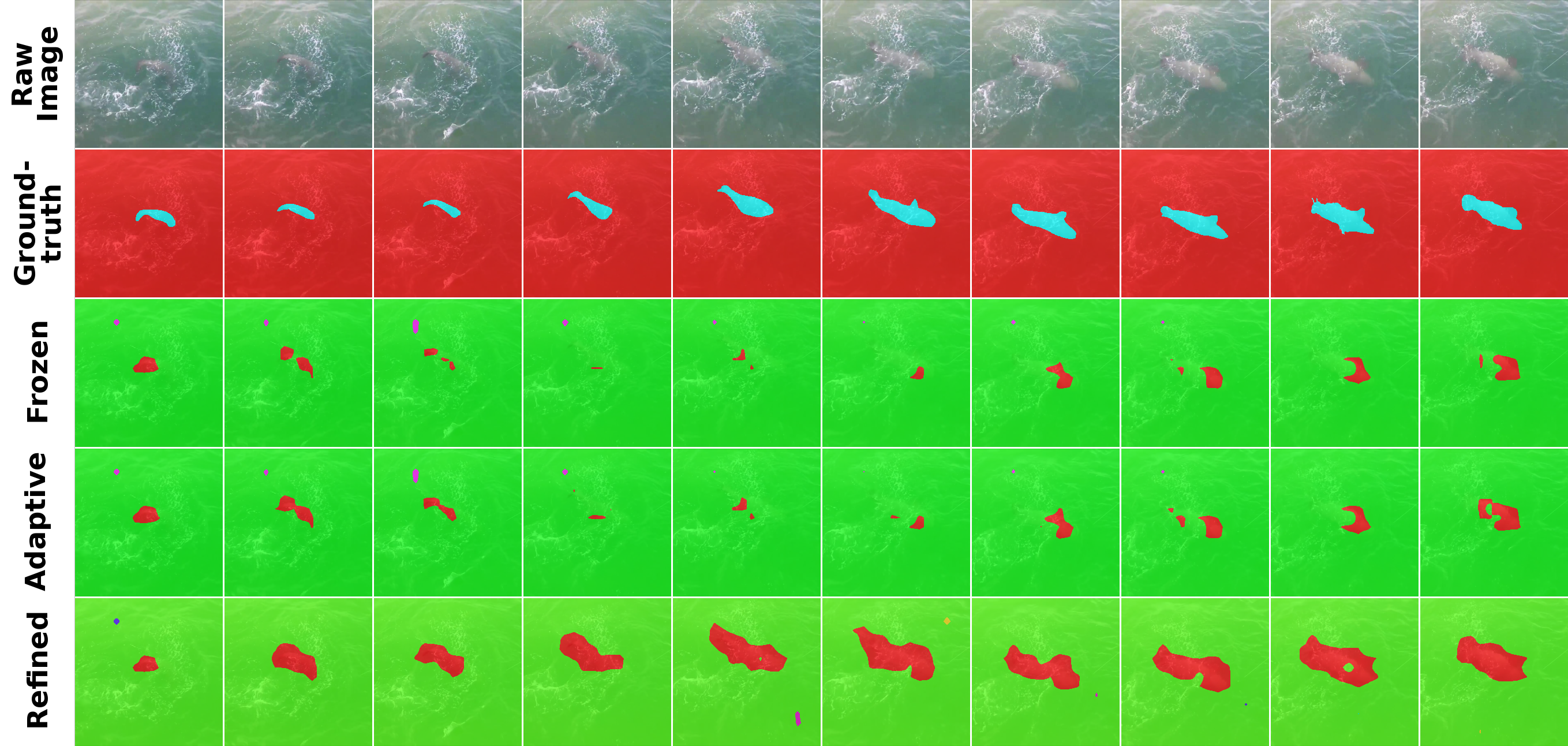}
    \caption{Dynamic scene recovery. Left: a new skateboarder enters at frame 4. The Frozen baseline fails by design, while the Adaptive mechanism detects the semantic shift and triggers an update, shown by the red border. Right: an object is partially occluded and blurry in the initial frames. The Adaptive mechanism does not trigger because the object is already present, whereas uniform refinement continuously improves slot quality and preserves accurate boundaries over time.}
    \label{fig:dynamic_scenes}
\end{figure}

If a significant proportion of patches exhibit high distance values $d_{t,i}$, the framework flags the frame as containing novel semantics and triggers a slot refresh (standard slot attention iterations). The core intuition of SemanticSlots-Adaptive is to dynamically detect when a new object enters the scene, updating the slots only on these flagged frames to save computation. For example, when a skateboarder enters the scene (Fig.~\ref{fig:dynamic_scenes}, left), its unmatched patch features trigger this semantic flag, forcing an instant slot update to capture the new entity.

\begin{table}[ht]
\centering
\caption{Inference efficiency vs. segmentation quality on YTVIS.}
\label{tab:adaptive_efficiency}
\small
\begin{tabular}{lccc}
\toprule
\textbf{Inference Mode} & \textbf{Time / Video (ms)} & \textbf{mBO} & \textbf{ARI} \\
\midrule
SemanticSlots-Frozen  & \textbf{58.6} & 60.8 & 86.4 \\
SemanticSlots-Adaptive & 60.8 & 61.3 & 86.6 \\
SemanticSlots-Refined  & 69.5 & \textbf{62.1} & \textbf{86.9} \\
\bottomrule
\end{tabular}
\end{table}

Figure~\ref{fig:dynamic_scenes} and Table~\ref{tab:adaptive_efficiency} show the trade-off. SemanticSlots-Adaptive recovers part of the gap between Frozen and Refined with little additional cost, improving mBO from 60.8 to 61.3 while increasing inference time only from 58.6 ms to 60.8 ms per video. However, the right example in Figure~\ref{fig:dynamic_scenes} shows that novelty-based triggering is limited: it can detect newly appearing semantics, but it may miss cases where an object is already present yet poorly represented due to blur, occlusion, or weak initial binding. This shows that slot refreshing each frame is not only useful for discovering new objects; it also improves existing slots over time. For this reason, SemanticSlots-Refined remains the most robust inference mode, reaching the best mBO and ARI in Table~\ref{tab:adaptive_efficiency}.

\paragraph{Cross-Attention to Frame Features.}
To isolate whether the benefit comes from the Transformer architecture itself or from cross-attention to frame features, we train a variant where the decoder has no access to feature context. Instead of using previous patch features autoregressively, the decoder uses only positional query embeddings, similar to DETR~\cite{carion2020end}, while retaining cross-attention to slots. Table~\ref{tab:context_ablation} shows the results on YTVIS dataset.

\begin{table}[ht]
\centering
\caption{Transformer decoder with and without access to frame features on YTVIS.}
\label{tab:context_ablation}
\small
\begin{tabular}{lcccc}
\toprule
\textbf{Decoder} & \textbf{Context} & \textbf{ARI} & \textbf{mBO} \\
\midrule
Transformer & \checkmark (features) & \textbf{86.6} & \textbf{62.8} \\
Transformer & -- (pos. queries) & 45.8 & 40.0 \\
\bottomrule
\end{tabular}
\end{table}

Without access to frame features, performance drops significantly, confirming that the cross-attention to features is the main advantage and not the Transformer architecture. 

\paragraph{Training Efficiency.}

A practical advantage of SemanticSlots is its vastly reduced training footprint. Standard video OCL methods process $T{=}6$ frames per iteration, requiring the ViT encoder to extract features for each frame, Slot Attention to update recurrently, and temporal losses to be computed. In contrast, SemanticSlots trains exclusively on single frames ($T{=}1$), eliminating this massive computational overhead. Moreover, because all three of our evaluated inference regimes (Frozen, Adaptive, and Refined) share the exact same underlying model parameters, they all directly inherit this training efficiency without requiring any video-level fine-tuning. Table~\ref{tab:efficiency} compares training time for 50k iterations on MOVi-C (batch size 64, single A100 GPU).
\begin{table}[ht]
\centering
\caption{Training sequence length and efficiency. Frames/iter denotes the number of frames processed per iteration. By training on single frames, SemanticSlots trains significantly faster than other methods.}
\label{tab:efficiency}
\small
\begin{tabular}{lccc}
\toprule
\textbf{Method} & \textbf{Frames/iter} & \textbf{Time (h)} & \textbf{Relative to SemanticSlots} \\
\midrule
VideoSAUR & 6 & $\sim$22 & 1.4$\times$ \\
SlotContrast & 6 & $\sim$24 & 1.5$\times$ \\
RandSF.Q & 6 & $\sim$26 & 1.7$\times$ \\
\textbf{SemanticSlots} & 1 & $\sim$16 & -- \\
\bottomrule
\end{tabular}
\end{table}

\paragraph{Fixed-Slot Stability and Position Invariance}
    To empirically validate position-invariance, we apply our strict SemanticSlots-Frozen inference regime to alternative decoder architectures. Slots are computed once from the first frame ($t{=}0$) and held constant for 36 subsequent frames, exactly as in our frozen mode. No temporal predictors, slot updates, or iterative refinements are applied; each decoder must rely solely on its internal mechanism to bind the fixed semantic signatures to moving object features.

\begin{figure}[t!]
    \centering
    \includegraphics[width= 0.6 \linewidth]{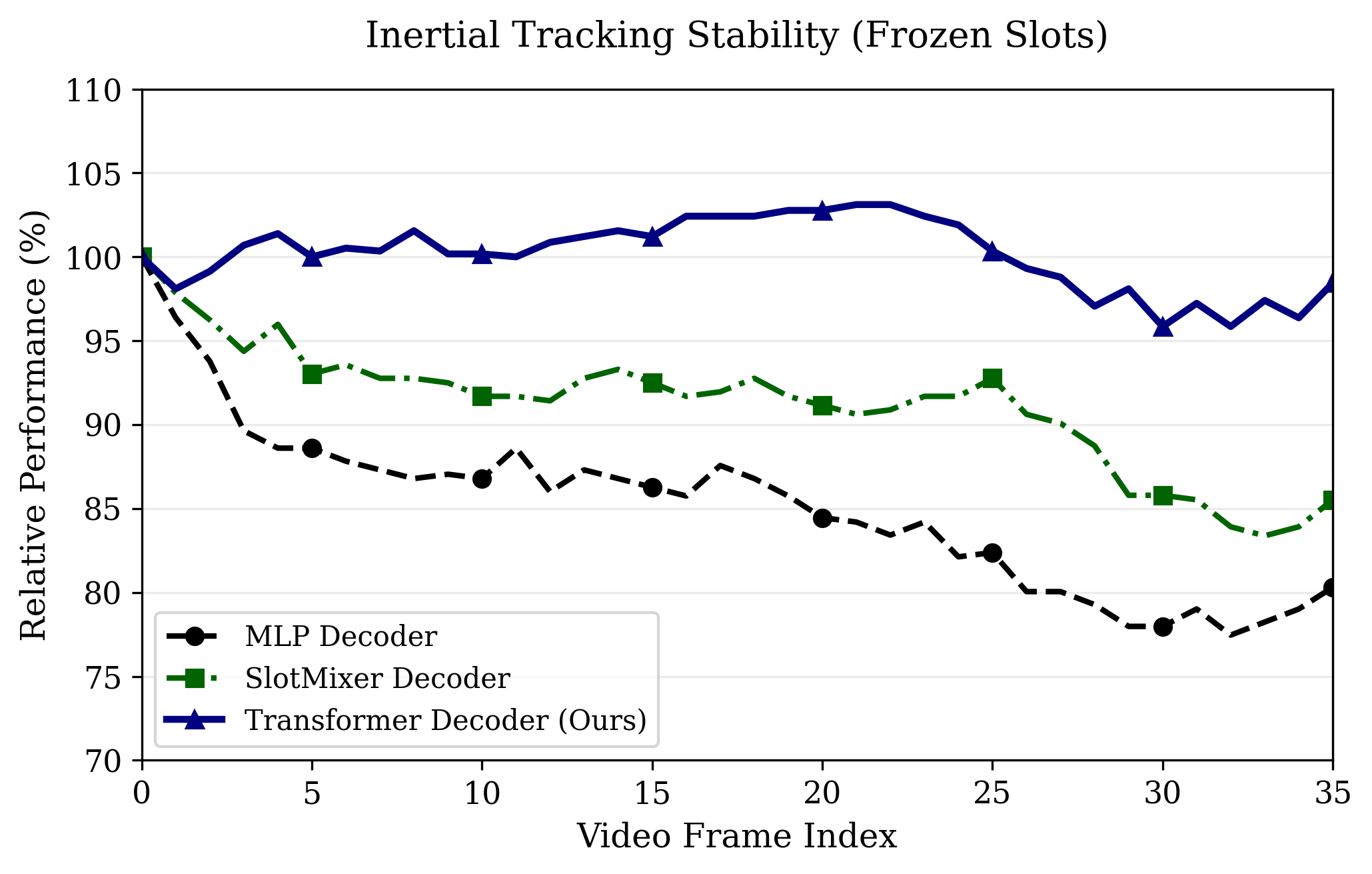}
    \caption{\textbf{Fixed-Slot Decomposition Stability.} Comparison of mBO over 36 frames using frozen slots from the first frame. The Transformer decoder (blue), with access to features from subsequent frames, shows negligible degradation, proving it successfully retrieves object identities regardless of motion. In contrast, ``blind'' decoders (black, green) fail to adapt to object displacement, demonstrating the fragility of coordinate-locked representations.}

    \label{fig:tracking_graph}
\end{figure}

As illustrated in Figure~\ref{fig:tracking_graph}, our Transformer-based decoder with cross-attention demonstrates a unique capacity for fixed-slot decomposition. While traditional ``blind'' decoders (MLP and SlotMixer) exhibit significant performance collapse, degrading by 19.5\% and 14.7\% respectively, the Transformer decoder maintains a nearly flat mBO curve with a negligible degradation of only 1.4\%. This 20-point performance gap relative to the very first frame confirms that traditional decoders are structurally unable to re-localize objects under motion.

\paragraph{Component Removal.}
To isolate the role of each temporal module, we adapt the video-trained VideoSAUR baseline ($T{=}6$) to use our Transformer decoder. Table~\ref{tab:ablation} shows the results of stripping these components versus moving to our image-only paradigm ($T{=}1$).

\begin{table}[t!]
\centering
\caption{Ablation study on YTVIS. We isolate the impact of temporal modeling components under video-level training versus our image-only training paradigm. ($^{*}$Denotes iterations applied strictly at inference time).}
\label{tab:ablation}
\small
\begin{tabular}{lcccccc}
\toprule
\textbf{Configuration} & \textbf{Train Data} & \textbf{Pred.} & \textbf{Tsim} & \textbf{Iter} & \textbf{mBO} & \textbf{ARI} \\
\midrule
VideoSAUR (baseline) & Video ($T{=}6$) & \checkmark & \checkmark & 2 & 59.8 & 81.2 \\
\quad $-$ Tsim loss  & Video ($T{=}6$) & \checkmark & --         & 2 & 59.6 & 81.7 \\
\quad $-$ Iterations & Video ($T{=}6$) & \checkmark & \checkmark & 0 & 47.8 & 69.9 \\
\quad $-$ Predictor  & Video ($T{=}6$) & --         & \checkmark & 2 & 43.7 & 49.3 \\
\quad Iter=1         & Video ($T{=}6$) & \checkmark & \checkmark & 1 & 42.8 & 54.8 \\
\midrule
\textbf{SemanticSlots-Frozen} & Image ($T{=}1$) & -- & -- & 0 & 60.8 & 84.1 \\
\textbf{SemanticSlots-Refined} & Image ($T{=}1$) & -- & -- & $3^{*}$ & \textbf{62.8} & \textbf{86.6} \\
\bottomrule
\end{tabular}
\end{table}

Table~\ref{tab:ablation} shows that within the video pipeline, removing individual components degrades performance. Yet, removing all temporal machinery and switching to image-level training achieves the best results. This counterintuitive finding suggests that these components, while designed for temporal consistency, complicate optimization when combined with a context-aware decoder. The Transformer decoder retrieval mechanism is sufficient, making additional video machinery entirely redundant. Reintroducing slot iterations optionally at inference simply serves as a lightweight refresh to further improve performance.

\paragraph{Discussion on Challenges and Future Work.} While our approach successfully handles object entry, occlusion, and motion blur, the framework prioritizes semantic retrieval over instance-level temporal continuity. This can cause ambiguity when distinguishing between identical entities in a scene. Future work should investigate how to re-integrate temporal priors into this retrieval-based framework to mirror the human ability to balance semantic identity with temporal persistence. We do not view this as a discouragement of temporal modeling, but rather as an invitation to develop effective ways of combining explicit motion priors with the powerful object-centric search capabilities demonstrated in this work.

\section{Conclusion}

In this paper, we identified a critical structural bottleneck in video OCL: the spatial entanglement inherent in context-free decoders. By shifting to a retrieval-based Transformer mechanism, we decoupled semantic identity from spatial location, showing that slots computed from a single frame are sufficient to decompose every subsequent frame of a video. This finding suggests that complex temporal predictors and auxiliary losses used by current video OCL methods compensate for a decoder limitation rather than addressing a fundamental video problem. Our method sets a new state-of-the-art performance on YouTube-VIS with 86.6\% ARI and 62.8\% mBO, achieving these results through architectural simplification rather than auxiliary temporal complexity.

\paragraph{Ethical Statement.} There are no ethical issues.

\section*{Acknowledgements}

This research was supported by the Fonds de recherche du Québec -- Nature et technologies (FRQNT) through the Team Research Project program (Grant No. 327313; DOI: 10.69777/327313). This work was also supported in part by the OPSIDIAN program at Polytechnique Montr\'eal, funded through the Natural Sciences and Engineering Research Council of Canada (NSERC) CREATE Program. Computational resources were provided by the Digital Research Alliance of Canada (formerly Compute Canada). The authors gratefully acknowledge this support.

\bibliography{egbib}
\end{document}